\pdfoutput=1

\documentclass[11pt]{article}

\usepackage[]{acl2022}
\usepackage{times}
\usepackage{latexsym}

\usepackage{microtype}

\usepackage{adjustbox}
\usepackage{amsmath,amssymb}
\usepackage{bm}
\usepackage{booktabs}
\usepackage[nameinlink]{cleveref}
\usepackage{dirtytalk}
\usepackage{enumitem}
\usepackage{subcaption}
\usepackage{textcomp}
\usepackage{tikz}
\usepackage{hyperref}

\usepackage[normalem]{ulem}

\usepackage[textsize=large]{todonotes}  

\title{An Item Response Theory Framework for Persuasion}

\author{Anastassia Kornilova  \And  Daniel Argyle   \\
  FiscalNote Research \\
  \texttt{anastassia, daniel, vlad@fiscalnote.com} 
  \And Vlad Eidelman 
  \\}

\date{}

\begin{document}
\maketitle
\begin{abstract}
In this paper, we apply Item Response Theory, popular in education and political science research, to the analysis of argument persuasiveness in language. We empirically evaluate the model's performance on three datasets, including a novel dataset in the area of political advocacy. We show the advantages of separating these components under several style and content representations, including evaluating the ability of the speaker embeddings generated by the model to parallel real-world observations about persuadability.


\end{abstract}


\section{Introduction}


Persuasion is the art of instilling in someone a given belief or desire to take a given action. 
The action can be expressing agreement with the speaker in a debate \citep{DDO_Users_2019}, making a donation to a crowdfunding campaign \citep{crowdfunding} or non-profit \citep{donation_2019}, or a Supreme Court ruling \citep{Danescu-Niculescu-Mizil+al:12a}. 
Social psychology frameworks for understanding persuasion, such as the Elaboration Likelihood Model (ELM), argue that attributes of successful persuasion fall into three groups: (1) message, the text of the argument; (2) audience; and (3) speaker, the source of the argument. \citep{petty1986elaboration, audience_effects_2017, cialdini2009influence}. 

Although much attention has been given to studying the text, text in isolation fails to capture how the audiences' prior beliefs and predispositions can affect their response to the same argument. Several recent studies have considered all three factors within the context of specific datasets by creating features to represent the audience as a whole or by building separate models for different types of audiences \citep{audience_effects_2017, CMV_Original, DDO_Users_2019, NYTIMES_2020}. In this paper, we present a broad framework that can represent individual audience members in one model across a diverse set of persuasion tasks.


Since implementing the ELM framework requires separate data about the speaker, audience, and argument, it is difficult to validate  empirically. Often, we only have access to the observed outcome (e.g.\ did the person donate money). Both the persuadability of the audience and the persuasiveness of the argument are unobserved. Motivated by this, we explicitly model a persuasive scenario as a function of latent variables describing the persuadability of the audience and the persuasiveness of the text.

Our approach is based on Item Response Theory (IRT), a framework for modeling the interaction between latent traits and observable outcomes. While these types of models are well known in the context of education \citep{fischer_linear_1973, lord1980applications,mccarthy-etal-2021-jump} and politics \citep{clinton2004statistical}, to our knowledge this is the first application of an IRT model to study persuasion. Using this framework, we model the interaction between the grouped \textit{argument} and \textit{speaker}, and the \textit{audience}, explicitly. The argument and speaker are grouped together because in practice it is hard to separate their effects, especially in the written tasks covered in this study.
 

We explore two variations on the IRT framework and apply it to three different persuasion tasks. 
In addition to two previously studied tasks, we introduce a novel setting related to political advocacy group campaigns, where a recipient is asked by an organization to take a specific action. 


We evaluate these models with different parameterizations, including style and content features, showing that they are both effective for predicting persuasion, and have the ability to uncover latent characteristics of the audience that were modeled explicitly in previous works. 





Our contributions are as follows: 1) we formalize the use of IRT model formulations for persuasion and show the advantages of them over existing approaches, 2) we introduce a new dataset of political advocacy emails, 3) we apply the formulations with style and content features on three persuasion tasks, and 4) we show that the separate latent audience component is interpretable and consistent with external information. All code associated with the paper is available at \url{https://github.com/akornilo/IRT_Persuasion}.




%



\section{Item Response Theory}

Item Response Theory (IRT) represents a set of models that explain an observed outcome based on latent traits. These models are frequently used when an outcome is easily observed, but the factors predicting that model are unobservable. For example,  in education an outcome could be a student's answer to an exam question, and the latent predictive traits are a students knowledge and the difficulty of the question; in politics an outcome could be a vote on a bill and the unobservable traits are the legislator's and bill's ideology. Crucially, an IRT model provides both a prediction of the outcome, and an interpretable measurement of the latent variables. 

In applying IRT to persuasiveness, we can view the audience as having a response to the item, where the item is an argument composed of the speaker and message pair.



\subsection{Rasch Testing Model}

We build on two specific IRT parameterizations. The first, the \textbf{Rasch} model \citep{rasch1960probabilistic} 
is commonly used in education research to model the difficulty of standardized test questions~\citep{fischer_linear_1973, lord1980applications}. In it the probability that an individual $i$ answers test question $j$ is given by:

\begin{equation}
    p(y_{ij}=1 \mid \alpha, \beta) = \sigma(\alpha_i - \beta_j)
    \label{eq:rasch}
\end{equation}

\noindent where $\alpha_i$ represents a respondent (e.g.\ a student's ability) and $\beta_j$ represents the item (e.g.\ the difficulty of a test question). Intuitively, if the ability is greater than the question difficulty, then the student will answer the question correctly. Given a series of exam sessions 
one can estimate values of $\alpha$ and $\beta$ for all of the students and questions in the dataset. This can be done using a variety of optimization strategies, such as Expectation Maximization or Bayesian techniques \citep{bock1981marginal, natesan2016bayesian}.

However, one limitation of this approach is that it cannot be used to perform inference on new test questions because all parameters are estimated simultaneously. To solve this problem, \citet{fischer_linear_1973} proposed the linear logistic test model that parameterizes the difficulty, $\beta$, as a weighted linear combination of test features. In this formulation, the student ($\alpha$) remains a latent variable, but the $\beta$ of an unseen question can be predicted using attributes of the question itself.

Following \citet{fischer_linear_1973}, the parameterization used to predict the item parameters is a weighted linear sum of features:
\begin{equation}
   \beta_j = \sum_{k=1}^K w_k \times \psi_{jk}
    \label{eq:2lmparam}
\end{equation}

\noindent where $\psi_k$ is an input feature representing the item, and $w_k$ is the associated weight.

In order to apply this model to persuasion, we propose considering argumentation as follows: First, arguments can vary in quality, similar to test questions having different difficulty levels. Second, we can only measure the quality of an argument based on how the audience reacted; 
similar to how a students ability is measured via their performance. 
Third, it is possible that a good argument is matched with an audience reticent to persuasion, 
similar to a good student receiving a particularly hard question. Note that this requires an audience member observe multiple arguments, and that each argument be heard by multiple audience members. Inspired by the linear logistic model, we model the latent argument parameter as a function of attributes of the argument itself, thus allowing us to include attributes of the speaker and text in the model directly.

\subsection{Two Parameter IRT}

While the simplicity of the Rasch model is powerful, a two parameter generalization of an IRT model (a two parameter logistic - \textbf{2PL}) provides additional benefits for our application \citep{twoparamlog}. In the simplest version, a two parameter model (so called because the item is modeled with\textbf{} two parameters) is as follows:
\begin{equation}
    p(y_{ij} = 1 \mid \bm{\alpha}, \bm{\phi}, \bm{\beta}) = \sigma\left(\alpha_i \cdot \phi_j - \beta_j\right)
    \label{eq:basic2p}
\end{equation}

\noindent where as before, $\alpha_i$ represents the respondent (students ability), and $\beta_j$ is the item's difficulty,\footnote{Analogous to the Rasch model, this tells us the overall difficulty level of the question} but now $\phi_j$ represents the item's discrimination.\footnote{Discrimination is how well the question is able to tell which students perform better, a high value indicates clearly separates high scoring students from low scoring, a negative value would indicate that low performing students are more likely to get the question right than high performing.} 
We similarly generalize this model by estimating the two item parameters, $\beta_j$ and $\phi_j$, as linear functions of features as in Equation \ref{eq:2lmparam}. 

This framework has commonly been be used to explain legislator voting behavior \cite{clinton2004statistical}, a useful analogy as many of the persuasion contexts we consider have political undertones. In this case, the response $y_{ij}$ is a vote by respondent $i$ (a legislator) on item $j$ (a bill). \citet{clinton2004statistical} show that the parameter $\alpha_i$ can then be interpreted as the respondent's ideology (e.g negative values are more liberal, positive values are more conservative); $\phi_j$ is referred to the bills polarity (i.e.\ discrimination);\footnote{Large negative or positive values indicate that a bill is strongly ideological, a value close to zero means the vote isn't strongly driven by ideology.} $\beta_j$ represents the bill's popularity (i.e.\ difficulty).\footnote{Large values indicate a bill that is ``difficult'' to vote for and is less likely regardless of ideology.} 
Persuasion is a generalization of this framework because popularity can correspond to properties of arguments that are appealing overall, while polarity represents techniques or topics that appeal only to a subset of the audience.

\subsection{Audience Analysis}

Once a Rasch or a 2PL model is fit, the learned $\alpha$ can be interpreted as a one-dimensional respondent embedding. In the legislator voting context these values can be interpreted as ideologies: legislators with very negative or very positive embeddings reflect very liberal and conservative stances, respectively, while those with small-value embeddings map to moderate legislators. While interpretation of these values will depend on the task, in general, similar embeddings will map to similar audience members and can be grouped together for further analysis.

\section{Related Works}
 
\textbf{Audience Effects} The properties of the audience in relation to argument persuasiveness have previously been examined in several predictive studies. \citet{audience_effects_2017} show that audiences with a more ``open'' personality respond better to emotional arguments, while \citet{NYTIMES_2020} show that liberals are more affected by the style of a new editorial than conservatives. \citet{donation_2019} also find that people with different personality types respond differently to emotional vs.\  logical appeals. \citet{CMV_Original} show how ``malleable'' different Reddit users are to new perspectives. \citet{DDO_Politics_2018, DDO_Users_2019} show that prior beliefs play a strong role in how persuadable someone is. \citet{cano-basave-he-2016-study} study persuasiveness of style in political speeches. In contrast to these studies, our method is designed to work when we have limited or no information about the audience of an argument.

\textbf{Item Response Theory} As described in the previous section, IRT models have primarily been applied in politics to measure the ideology of politicians \citep{clinton2004statistical, poole1985spatial}. While most IRT implementations here rely only on the responses as data, more recent work augment the models to take advantage of the text through a simultaneously estimated topic model~\cite{gerrish2012issueadjusted, vafa2020text, lauderdale_clark.2014}.

The efficacy of IRT has been applied on large-scale datasets to verify the validity of standardized tests both in the U.S. and internationally \citep{aera.book14, irt_handbook}. Recent advances have focused on polytomous test questions and creating new questions \citep[the `cold-start' problem:][]{settles_2020_ml, mccarthy-etal-2021-jump}. In this paper, we focus on the simplest form, but this area of research points to many possible extensions.

\textbf{Argument Quality} Argument mining has been studied in various domains~\citep{palau2009argumentation}. Most relevant here, several studies have attempted to study argument quality through pairwise ranking as the outcome \citep{habernal-gurevych-2016-argument, DBLP:journals/corr/abs-1907-08971, DBLP:journals/corr/abs-1909-01007}. 

\textbf{Framing Theory} In the study of framing effects, the expectancy value model~\citep{framing_theory} represents an attitude as $\sum_i v_i \times w_i$, where $v_i$ is the favorability of the object of evaluation (e.g.\ a candidate), on dimension $i$ (e.g.\ foreign affairs or personality), and $w_i$ is the salience weight ($\sum_i w_i=1$). Our parameterization of $\beta_j$ and $\phi_j$ can be seen in this paradigm as identifying frames in communication, with each feature of the style and content as a dimension, and learning the framing effect of each. 


\section{Datasets}

In order to apply the IRT framework, an audience member must respond to multiple arguments (and arguments must be observed by multiple audience members). Too few responses implies that an audience member's latent value will be driven entirely by the one or two arguments. While not many existing argument mining datasets meet this criteria, we are able to study three diverse settings. 
Additionally, our advocacy task is akin to many real-world settings where users on one-platform are asked to complete an arbitrary task (e.g.\ a retail mailing list getting users to click on a promotion).

\subsection{NYTimes Editorials}

The NYTimes Editorial corpus\footnote{\url{https://webis.de/data/webis-editorial-quality-18.html}} consists of 975 editorials from the New York Times news portal \citep{NYTIMES_2018}. Each publication was reviewed by 3 conservatives and 3 liberals from a pool of 12 conservative and 12 liberal reviewers.

Each reviewer rated the editorials as either `challenging', `reinforcing' or `no effect'. These labels must be approached with care as reinforcing could imply `reinforced view against the article's stance'. \citet{NYTIMES_2020} study this corpus in a ternary setting by aggregating the liberal and conservative votes and building separate models for each side. For our study, we construct a binary task for predicting `whether this article had an effect'. While this framing elides whether the speaker succeeded according to her intent, it does relay whether the argument was persuasive. 

\subsection{Debates (DDO) Corpus}

\textbf{DDO} is a corpus of 78k debates scraped from \url{debate.org}.\footnote{\url{https://www.cs.cornell.edu/~esindurmus/ddo.html}} Each debate has two speakers and an audience votes on a winner.\footnote{While the audience can assign points to various aspects of the debate, this study will only consider the cumulative sum of the points.} In addition, each audience member can fill out their profile with their political and religious ideology, and stance on various political issues (e.g.\ Abortion or the Border Wall). Originally, it was used to study how prior beliefs and  similarities between the audience and the speaker affected debate outcomes \citep{DDO_Politics_2018, DDO_Users_2019}.

To preprocess the data, we removed all debates that have fewer than three rounds, end in a forfeit or a tie, have fewer than 100 words per side, or have fewer than 5 points awarded total. In addition, we excluded debates not on the following issues: Politics, Religion, Society, Philosophy, Education and Economics. Since we are interested in modeling individual audience members, we identify audience members who have responded on at least 10 debates, then remove debates where none of those members responded. The final dataset contains approximately 60k datapoints; 6320 debates and 1131 responders.  

Each debate has one side with a pro argument and one side with a con argument, resulting in the wining side being ``assigned more points''. The prediction task consists of whether a responder gave more points to a given debate side. Since our models only consider one argument at a time, we treat each side of the debate as a separate item, concatenating texts from all rounds from that speaker.\footnote{We are interested in how a single unit of argument affects the audience, and leave extension of this to account for both simultaneously to future work.}



\subsection{Advocacy Campaign Corpus}

Grassroots advocacy is the process wherein organizations (e.g.\ companies, non-profits, coalitions) encourage individual citizens to influence their government. In the United States, such lobbying often takes the form of advocacy email campaigns, sent by an organization to specific audiences, asking them to take an action, such as contacting their legislators to vote yes or no on a particular bill. 


We construct a dataset containing the text and metadata of these emails, from a popular advocacy software platform, paired with whether recipients took the requested action.\footnote{Due to privacy concerns, this dataset will not be released, but platform users agreed to terms of services providing for internal analysis.} 
Organizations will send different messages to the same audience over time, allowing us to identify which emails (items) elicited a response from specific recipients. Thus, it is possible to distinguish messages that did not generate interest overall (popularity) from messages that did not resonate with specific groups of recipients (polarity). 

The dataset contains 63,795 individual recipients of 7,067 email campaigns from 328 different organization, resulting in approximately 2 million individual data points. Each recipient has data for 15 to 100 emails and had an action rate between of 5\% - 95\%.\footnote{Those with a lower or higher action rate are unlikely to be illustrative of persuasion characteristics.} Each email included in the dataset had at least 6 responses. 

The data is not balanced with respect to organizations; while the largest organizations sent over 200 emails, the median is 6. 
One possibility of this imbalance is overfitting a feature that is only pertinent to one, particularly prevalent organization. To mitigate such effects, we include an indicator variable to specify the organization.\footnote{Alternatively, we could construct separate models for each organization, but refrain from doing so for three reason. First, about a quarter of recipients are `multi-org' - they receive emails from multiple sources, thus, we would like to model their behavior across all of them. Second, as many of the organizations are not well represented, they benefit from patterns that appear across different organizations. Finally, maintaining a separate model for every recipient and recipient is not as efficient or scalable.}

\section{Model Features}
\label{section:features}

Argument analysis is often separated into {\it style} and {\it content} features \citep{cano-basave-he-2016-study, ddo_undecided, NYTIMES_2020}, with additional categories included for argument quality and task specific properties. Since we group the speaker and the argument text together, we combine features representing both as inputs to $\phi$ and $\beta$. 



\paragraph{Lexicon Style Features} Style features represent higher-level properties of words and rhetorical structures. We chose the following sets of such features from lexicons that were commonly used in previous argumentation literature:

LIWC lexicon of 93 metrics ranging from parts-of-speech to thinking styles to emotions \citep{liwc};\footnote{We purchased a copy of the software from \url{liwc.wpengine.com} to obtain these labels.} 
Valence, Arousal, Dominance \citep{valence}; Concreteness \citep{concrete}. (These features were shown to be useful for argument quality analysis by \citet{CMV_Original}.)
Argument features developed by \citet{mpqa_arguing}, including \textit{necessity, emphasizing, desire, contrasting} and \textit{rhetorical question}; 
NRC Lexicon: Word-level level associations for emotions like anger, disgust and fear \citep{nrc_lexicon};
Sentiment and Subjectivity: as implemented in the TextBlob Python Library.\footnote{\url{https://textblob.readthedocs.io/}}

\paragraph{Argument Text} 
We use TF-IDF unigrams to represent the text directly (tuned with respect each task). While we initially explored using deep, contextual text representations, they did not show benefit, and the motivation for this paper is to understand the benefits of the IRT framework, rather than optimize performance based on the argument alone.

\paragraph{Debate-Only Speaker Features} In the debate platform, users can optionally specify a stance - for, against, undecided or no stance - on 48 issues such as Abortion, Death Penalty or Gay Marriage. These can be viewed as a proxy for the content as users often present arguments that align with their views.

\paragraph{Advocacy-Only Org Indicator} An indicator to account for the large variation in action rate between organizations. Additional indicators are used to represent the industry and organization size.

\paragraph{Advocacy-Only Appeals} Using data from \citet{donation_2019}, we built a multi-class classifier to recognize `emotional', `logical' and `credibility' appeals. The classifier was applied at a sentence level to the emails, and features were created for the average and the sum of the scores across the sentences.

\paragraph{Advocacy-Only Misc Features}: The day of the week and time of day have a strong effect on email click rate.\footnote{\url{https://sleeknote.com/blog/best-time-to-send-email}} We include indicator features for the day of the week and the hour of day. We include an \textit{urgency} indicator feature, based on a custom list of words indicative of high urgency and timeliness (e.g.\ ``soon'', ``now'', ``hurry'').

\paragraph{IBM Quality} \citet{ibm_large_quality} released a dataset of 30k sentence-level arguments with 0-1 quality ratings. Unlike our tasks where quality is a latent property, these sentences were assessed for quality directly. We re-implemented the BERT-FT model from this paper, using the MACE-P score. 
Since these scores were trained on short texts, we apply them to individual sentences in the input text, then use the min, max, average, range, 25th, 50th, and 75th percentiles of these scores. As far as we know, this is the first study to transfer the quality model to longer texts. These features will be grouped with Style for the analysis.

\section{Models and Results}
\label{sec:results}

Since the Editorials corpus is the smallest, we use the simpler Rasch parameterization, while the 2PL model is used for the Debates and Advocacy tasks.\footnote{In addition, there is natural polarity in the Debate task that lends itself to the 2PL model, as $\phi$ in equation \ref{eq:2lmparam} is designed to model such an effect.} Each of the models is trained using a regularized binary cross-entropy loss:
\begin{equation*}
\begin{split}
    L\left(\hat{y}_i, y_i\right) = - y_i \log \hat{y}_i  - (1 - y_i) \log \left(1 - \hat{y}_i\right)  \\ + c \cdot \|\alpha, \beta, \phi \| 
\end{split}
    \label{eq:loss}
\end{equation*}

\noindent where $\hat{y}_i$ is the output from equation \ref{eq:rasch} or \ref{eq:basic2p}, and $y_i$ is the binary label, representing if the persuasion was successful. The second part of the equation represents a regularization parameter. Details on the experimental parameters can be found in Appendix \ref{appendix:model}. For each task, an audience prior baseline is used. It is generated by calculating the rate at which the audience member was persuaded in the training data (e.g.\ did the article have an effect, how many recipients took the requested action), then drawing labels on the test data accordingly.


\subsection{Editorial Results}

The results on the Editorial task are shown in Table \ref{table:editorial}.  
The performance for all three feature sets is relatively similar, with all outperforming the audience prior. 

\begin{table}
\centering
\begin{tabular}{lr}
\toprule
Model    & Accuracy \\ \midrule
Audience Prior    & 0.662    \\ 
Style        &  0.741   \\
Text  &  \textbf{0.754}   \\
Style + Text & 0.750    \\ 
\bottomrule
\end{tabular}
\caption{Results for the Editorials Task (Rasch Model).}
\label{table:editorial}
\end{table}

\begin{figure}
    \centering
    \begin{subfigure}[b]{\linewidth}
        \includegraphics[width=\textwidth]{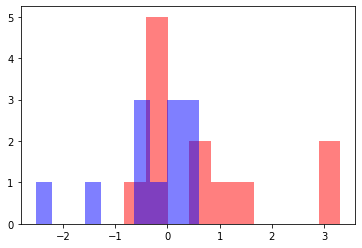}
    \end{subfigure}
    \caption{Reviewer Embeddings for the Editorial Rasch Model on the x-axis. Blue represents liberal reviewers, red represents conservative reviewers.}
        \label{fig:editorial_alphas}
\end{figure}

The embeddings and weights generated by the model can be analyzed separately for further insights. First, in Figure \ref{fig:editorial_alphas} we compare the distribution of audience embeddings ($\alpha$) for the liberal and conservative reviewers. According to our theory, these can be interpreted as individuals reticence to being persuaded. While a majority of reviewers are close to 0, we see two liberals with larger negative values (meaning they are particularly open to the messages) and several conservatives on the right (suggesting they are more closed off to these messages). This supports \citet{NYTIMES_2020} observation that conservatives are generally resistant to the {\it New York Times style}; however, the fact that the majority of reviewers from both sides have similar embeddings, suggests that the pattern is not very strong.

 This data also contained information from each reviewers Big 5 Personality test. We measured the Pearson correlation between the reviewers embeddings and found a strong correlation with extroversion (r=-0.568, p\textless 0.05) and openness (r=-0.344, p\textless 0.1). These findings closely match \citet{NYTIMES_2018}'s analysis between Big 5 Personality Ratings and the affectedness labels. The audience embedding is a latent parameter, thus, it does not explicitly represent personality or political preferences. This analysis has two implications: first, the IRT framework is successful in situations where additional data about the audience is not available; second, analyzing the embeddings lets us learn qualities of the audience post-hoc.

For style, the highly weighted features included negative sentiment markers (\textit{nrc\_negative, liwc\_negative\_emotions}); this aligns with \citet{NYTIMES_2020}'s observation that ineffective editorials tend to have a neutral tone (although their study only focuses on liberal reviewers). The quality features do not show consistent behavior: the \textit{quality\_mean} feature has a large negative weight (e.g sign of a bad editorial), but the 75th and 25th percentile features have positive weights; suggesting that the quality measure does not transfer well to editorials. 

\subsection{Debate Results}
\label{sec:res_debates}

The Debates data is approximately 10 times larger than Editorials and contains a more diverse audience. 
The results are shown in Table \ref{table:idealres}. 
Without the popularity parameter, $\beta$ the performance decreases, which confirms the theory that both polarity and popularity are necessary to adequately represent the argument and the speaker. The Speaker stance model outperforms just Text; a probable explanation is that the stances are a proxy for the actual opinions expressed in the text that a simple unigram representation can not capture.

\begin{figure}
    \centering
    \begin{subfigure}[b]{\linewidth}
        \includegraphics[width=\textwidth]{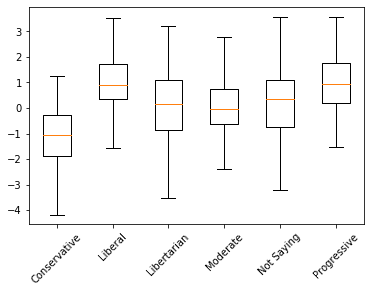}
    \end{subfigure}
    \caption{Distribution of one-dimensional audience embeddings on the y-axis.}
        \label{fig:ideal_1D}
\end{figure}

To understand the latent audience embeddings 
we compare them to the self-reported political affiliations from their profiles. Figure \ref{fig:ideal_1D} shows a clear separation between liberals and conservatives (the two largest groups). This finding supports the work of \citet{DDO_Users_2019} which showed that similarity on `Big Issue Stance' between the speaker and the audience member is a good indicator for predicting outcome. As with Editorials, the advantage of our approach is that we were able to infer audience member preferences without using their profiles. 

\begin{table}[t]
\centering
\begin{tabular}{lr}
\toprule
Model          & Accuracy             \\ \midrule
Random & 0.500 \\ 
Style & 0.561 \\
Text & 0.581  \\
Speaker & 0.611 \\
Speaker + Style  & 0.626 \\
 -$\beta$ (popularity) layer & 0.604 \\
\bottomrule
\end{tabular}
\caption{Results for Debates Task (2PL Model).}
\label{table:idealres}
\end{table}

To understand what $\phi$ and $\beta$ tells us about persuasive theory, we will focus on the Speaker+Style model:

\noindent \textbf{High Polarity}: Abortion, Gay Marriage, Progressive Tax;

\noindent \textbf{Low Polarity}: Border Fence, Gun Rights, Homeschooling;

\noindent \textbf{High Popularity}: quality\_max, quality\_range, liwc\_differ;

\noindent \textbf{Low Popularity}: liwc\_Exclam, liwc\_authentic, liwc\_drives

For popularity the significant factors are related to style and quality. The high `quality\_max' feature suggests that the quality model transfers better to this context than Editorials. 
The low popularity value for `liwc\_authentic' is interesting, as \citet{NYTIMES_2020} also found that authenticity generally led to No Effect editorials.

For polarity, the highest weighted are the stances. `Polarity High' corresponds to having a Pro stance on those issues, which in this case represent a Liberal view point. This corresponds with the Liberal recipient embeddings in Figure \ref{fig:ideal_1D} having generally positive embeddings (alignment in weights results in  positive final weight). The opposite is true for the Conservative issues and embeddings. This alignment reinforces the finding that prior beliefs play a strong role in outcomes \cite{DDO_Politics_2018}.

\begin{figure}
    \centering
    \begin{subfigure}[b]{\linewidth}
        \includegraphics[width=\textwidth]{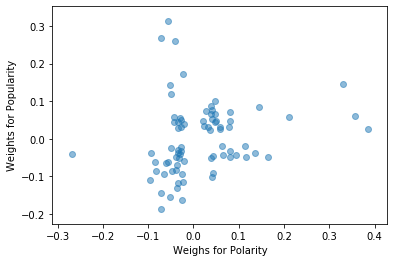}
    \end{subfigure}
    \caption{Contrast of weights from popularity vs polarity features.}
        \label{fig:pol_v_pop}
\end{figure}

Figure \ref{fig:pol_v_pop} plots the weights learned for each feature for the polarity and popularity parameters.\footnote{This figure excludes features that had very small weights along both dimensions.} 

Notably, the orthogonal pattern extends beyond the top features, features that strongly predict whether the audience responds to an argument do not usually strongly predict whether the argument is popular overall.




\begin{table*}[h]
\centering
\begin{tabular}{l rr rr rr}
\toprule
 & \multicolumn{2}{c}{Overall} & \multicolumn{2}{c}{Audience Average} & \multicolumn{2}{c}{Org Average} \\
 \cmidrule(lr){2-3} \cmidrule(lr){4-5} \cmidrule(lr){6-7}
                & Acc. & Macro-F1 & Acc. & Macro-F1 & Acc. & Macro-F1\\ \midrule
Org Prior      &  0.608 & 0.514 & 0.606 & 0.263 & 0.630 & 0.513  \\ 
Audience Prior   & 0.710 & 0.415 & 0.716 & 0.318 & 0.714 & 0.472  \\ \addlinespace

Org Only  & 0.757    & 0.667  & 0.759 & 0.589 & 0.728 & 0.573   \\ 
Org + Style & \textbf{0.781}  & \textbf{0.708} & \textbf{0.761}  & \textbf{0.662} & \textbf{0.771} & \textbf{0.678}    \\
  - $\beta$ (popularity) & 0.750 & 0.653 &  0.749 & 0.643 & 0.756 & 0.654  \\ 
\addlinespace
Sep Feat V1  &  0.725 & 0.619 & 0.726 & 0.571 & 0.700 & 0.520  \\
Sep Feat V2  &   0.748 & 0.678 & 0.750  & 0.604  & 0.698 & 0.654   \\
\bottomrule
\end{tabular}
\caption{Results For Advocacy Task (2PL Model).}
\label{table:advocacy_res}
\end{table*}
\subsection{Advocacy Results}


Table \ref{table:advocacy_res} shows the results for the Advocacy task.\footnote{Due to computational constraints, we omitted the raw text model from this task.}
The overall accuracy and macro-F1 scores represent results across all data, while the Org and Audience average accuracy represent data for individual organizations and respondents. Due to the variation in action rate and sample size, the macro-F1 results are particularly important.

While the Org Only model performs well,\footnote{One likely explanation for this performance is that audience is not independent of the speaker - by virtue of receiving emails from this organization, recipients may also have similar preferences.} 
the improved performance with the additional of Style suggests that the style of an email still affects the user. The style features may have an advantage for recipients associated with a diverse set of organizations. Without $\beta$, the performance is significantly worse, again confirming the need for both parameters. 

To better understand the effect of style and org features, two additional models are trained that separate between polarity and popularity. In \textbf{Sep Feat V1}, $\phi$ receives style features, $\beta$ receives org indicators. In this setting, ($\alpha\cdot\phi$) represents how individuals are affected by style, while $\beta$ models the organizations base rate. In \textbf{Sep Feat V2} the features are reversed. V1 has the worst performance of all five 2PL models, suggesting that modeling the interaction between the recipient and organization ($\alpha \cdot \phi$) is important. Org-Only and V2 have mixed performance on accuracy, but V2 performs better on macro-F1, suggesting that style influences the recipients' decisions to act.

Finally, we analyze the features with lowest and highest magnitudes from $\beta$ in the Org+Style model. The highest weighted features include \textit{concreteness, average-logical-appeal, word count} and \textit{quality 75th percentile}. The lowest weighted features (unlikely to produce action) include \textit{valence, quality mean, arousal} and \textit{liwc-we}. Similar to the Editorials, the quality features are contradictory, suggesting the connection between sentence level and document level quality needs to be investigated further. The logical appeal feature shows they are particularly effective (the corresponding scores for emotional and credibility appeals had smaller, negative weights).

\section{Conclusion and Future Work}
In this paper, we validate the social psychology frameworks for persuasion using the IRT framework to explicitly model the audience and the speaker. Our approach lets us analyze how different audience members respond to the same argument, and we show that our representation implicitly learns latent audience features modeled explicitly by other models. 

We empirically showed several additional insights about persuasion. In the Debates and Advocacy tasks, the Popularity parameter improved performance showing that certain stylistic elements are universally appealing. In the Debates task, the audiences' embeddings aligned with their political affiliation, showing that prior beliefs play a strong role in their argument perception. While the background information about the audiences was available for these tasks, we did not need to model it explicitly; as a result this setup allows us to make predictions for audiences who do not report their affiliation. 

A potential negative side of the models is they may learn latent characteristics of the speaker or audience they may not be aware of or consider private. However, all datasets studied in this paper were either public and anonymous or private with audiences who consented to analysis.

This study focused on simple representations to show the viability of our method and provide for explainability. To build on this foundation in future work, we will: expand argument text representations with contextual word embeddings and stance detection models; include higher dimensional embedding for audience and item parameters (the IRT models easily generalize to this set-up). These improvements will allow us to better capture the elements of persuasion, especially in a complex case like Advocacy.

\bibliographystyle{acl2022}
\bibliography{acl2022}

\appendix
\section{Model Training Details}
\label{appendix:model}

The models described in section \ref{sec:results} were trained as follows. In equation \eqref{eq:loss}, $c$ is set to $1e^{-4}$ for all experiments. $L2$ loss is used for the Editorials and Advocacy corpus and text model for Debates, $L1$ is used for the remaining models in the Debates corpus. Editorial models are trained for 200 epochs; Debates for 25; Advocacy for 5. A learning rate of $0.01$ is used for Editorials and Debates; $0.005$ is used for Advocacy.

All results are reported over 5-fold cross-validation, with the splits performed at an argument level. 
All models are fit using the AdamW optimizer. 
The $\alpha$ embedding initializations are drawn from a uniform distribution of $-0.5$ to $0.5$.

\end{document}